\documentclass[letterpaper, 10 pt, conference]{ieeeconf}  

\IEEEoverridecommandlockouts                              

\usepackage{epsfig}
\usepackage{multirow}
\usepackage{makecell}
\usepackage{amsmath}
\usepackage{amssymb}
\usepackage{times}
\usepackage{pifont}

\usepackage[ruled,vlined]{algorithm2e}

\usepackage{xcolor}
\usepackage{esvect}

\newcommand\eg{\textit{e.g.,}}
\newcommand\etal{\textit{et al.}}
\usepackage{multirow}

\title{\LARGE \bf
Motion Prediction in Visual Object Tracking
}

\author{Jianren Wang$^{*}$, Yihui He$^{*}$
\thanks{* indicates equal contribution. Jianren Wang, Yihui He are with the Robotics Institute, Carnegie Mellon University, 5000 Forbes Ave., Pittsburgh, PA 15213, USA
        {\tt\small \{jianrenw@andrew,he2@alumni\}@andrew.cmu.edu}}%
}

\begin{document}

\maketitle
\thispagestyle{empty}
\pagestyle{empty}

\begin{abstract}

Visual object tracking (VOT) is an essential component for many applications, such as autonomous driving or assistive robotics. However, recent works tend to develop accurate systems based on more computationally expensive feature extractors for better instance matching. In contrast, this work addresses the importance of motion prediction in VOT. We use an off-the-shelf object detector to obtain instance bounding boxes. Then, a combination of camera motion decouple and Kalman filter is used for state estimation. Although our baseline system is a straightforward combination of standard methods, we obtain state-of-the-art results. Our method establishes new state-of-the-art performance on VOT (VOT-2016 and VOT-2018). Our proposed method improves the EAO on VOT-2016 from 0.472 of prior art to 0.505, from 0.410 to 0.431 on VOT-2018. To show the generalizability, we also test our method on video object segmentation (VOS: DAVIS-2016 and DAVIS-2017) and observe consistent improvement. 

\end{abstract}

\section{Introduction}
\begin{figure*}
    \begin{center}
    \includegraphics[width=\textwidth]{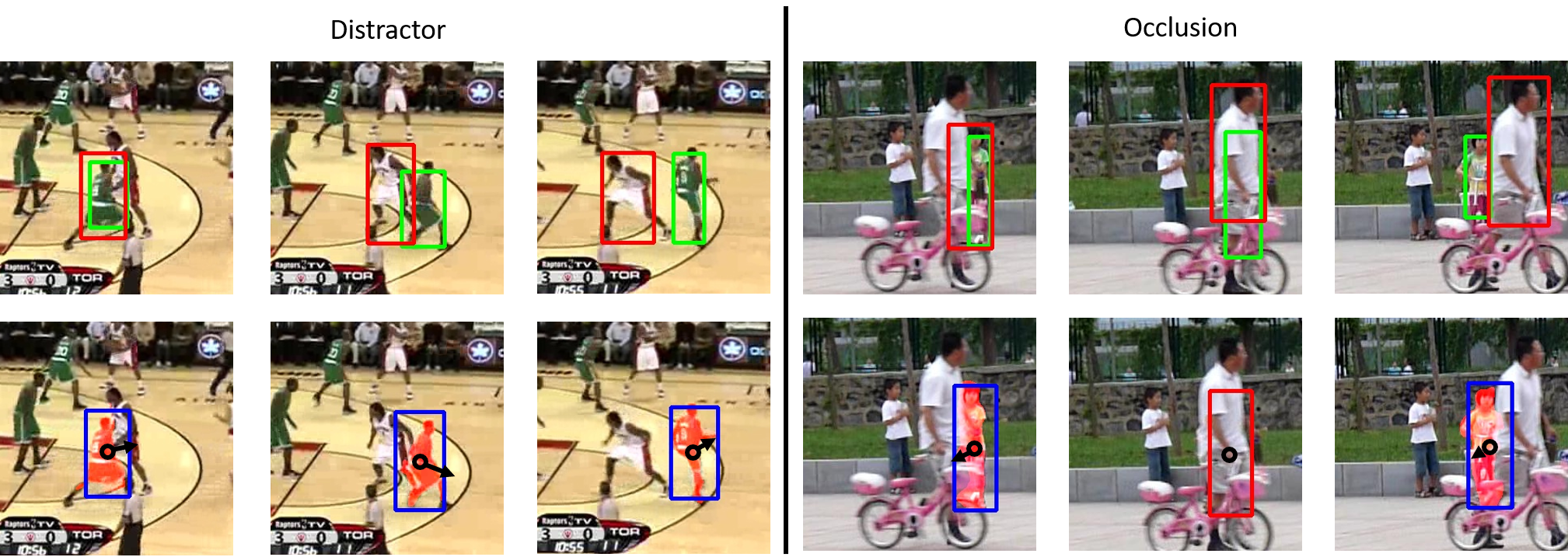}
    \caption{The first row shows the results of state-of-the-art tracker SiamMask~\cite{wang2018fast} (red) and ground truth (green). The second row shows the motion prediction (black arrow) and tracking results (with the blue bounding box and red segmentation mask) of our model. Our method improves the robustness against distractors and occlusions. (better view with color)}
    \label{fig:teaser}
    \end{center}
\end{figure*}

Tracking moving objects over space and time is fundamental for understanding the dynamic visual world, which has many practical applications in video processing, such as self-driving~\cite{lee2015road}, video surveillance~\cite{tang2017multiple}, and UAV navigation~\cite{mueller2016benchmark}. 

Many attempts have been addressed to improve the performance of trackers over the years. In the early days, motion model was a core component of tracking - constant velocity models~\cite{126006, 4103555}, Kalman filters~\cite{10.1007/3-540-55426-2_49, 489052}, particle filters~\cite{1415381,937594} and even social force models~\cite{5459260, johansson2007specification} for more complicated motions. In fact, in the early days, it was the dominant component, because (i) decent appearance descriptors were not available, and (ii) it had its roots outside of computer vision (\eg, tracking point targets in RADAR data), where there is no appearance information. However, most modern trackers assuming zero-velocity model, because (i) Modern tracking datasets contain many sequences with random camera motion, which fails most motion model~\cite{10.1007/978-3-319-48881-3_54, Kristan2018a} (ii) With better feature extraction and bounding box regression ability introduced by CNN~\cite{he2016deep, ren2015faster}, modern trackers~\cite{li2018high, li2018siamrpn++} can rely less on motion priors, which is known as tracking by detection. 

In contrast to prior works which tend to develop accurate systems based on more computational costly feature extractors, this work aims to develop a robust motion prediction method. We address the importance of motion prediction, even if trackers based only on appearance cues have already achieved good performance. We prove that even modern CNN trackers can benefit a lot from accurate motion predictions.

Concretely, in the first stage, we decouple camera motion and object motion. Second, we predict the object state in the future frame and create an adaptive search region for the detector to process. The adaptive search region focuses on smaller local regions when objects have slower speeds and smaller sizes, and vice versa. We then project the predicted state and search region back to the camera coordinate of the frame. We finally update the object state based on the measurement from the off-the-shelf object detector. Both state prediction and update are based on Kalman filter.

Our method has several benefits: First, by decoupling object motion from camera motion, we alleviate the motion noise caused by camera shake. Second, we free modern trackers from using only appearance information. As most tracking and segmentation methods can only discriminate foreground from the non-semantic background~\cite{zhu2018distractor}, the performance suffers significantly when the target object is surrounded by similar objects (know as distractors~\cite{zhu2018distractor}). Our method can also improve the performance under occlusions since the motion model can prevent the detector from tracking the occluders. We show in the experiments that our method improves the tracking performance by a large margin under both cases. We visualize part of the results in Fig.~\ref{fig:teaser}. Third, we achieve robust bounding box prediction by updating the state through the Kalman filter.

We evaluate our framework on major tracking datasets: VOT-2016~\cite{10.1007/978-3-319-48881-3_54} and VOT-2018~\cite{Kristan2018a}. We demonstrate the effectiveness of our method, both qualitatively and quantitatively. On VOT-2016, we achieve 0.505 EAO, and on VOT-2018 we achieve 0.431 EAO. Although our method focuses on video object tracking, it generalizes well to video object segmentation. Consistent improvements over the baseline method are demonstrated on VOS (DAVIS-2016~\cite{perazzi2016benchmark} and DAVIS-2017~\cite{pont20172017}). 

We summarize our contributions as follows: First, we revisit motion prediction in visual object tracking, which has long been ignored. Second, we propose a method that combines motion decouple, motion prediction with off-the-shelf appearance-based trackers. Third, our proposed method achieves state-of-the-art performance on VOT and can also consistently improve the performance on VOS.

\section{Related Works}
\subsection{Video Object Tracking}

In tracking community, significant attention has been paid to discriminative correlation filters (DCF) based methods~\cite{bolme2010visual,ma2015long,li2014scale,galoogahi2017learning}. These methods allow discriminating between the template of an arbitrary target and its 2D translations at a breakneck speed. MOSSE ~\cite{bolme2010visual} is the pioneering work which proposes a fast correlation tracker by minimizing the squared error. Performance of DCF-based trackers has then been notably improved through the using of multi-channel features~\cite{henriques2015high,danelljan2014adaptive,kiani2013multi}, robust scale estimation~\cite{danelljan2014accurate,danelljan2017discriminative}, reducing boundary effects~\cite{danelljan2015learning,kiani2015correlation} and fusing multi-resolution features in the continuous spatial domain~\cite{danelljan2016beyond}.

Tracking through Siamese Network is also an important approach~\cite{koch2015siamese,tao2016siamese,bertinetto2016fully,valmadre2017end}. Instead of learning a discriminative classifier online, the idea is to train a deep siamese similarity function offline on pairs of video frames. At test time, this function is used to search for a candidate similar to the template given in the starting frame on a new video, once per frame. GOTURN~\cite{held2016learning} used a deep regression network to predict the motion between successive frames. SiamFC ~\cite{bertinetto2016fully} implemented a fully convolutional network to produce a correlation response map with high values at target locations, which establishes a basic form of modern Siamese framework. Many following works have been proposed to improve the accuracy while maintaining fast inference speed by introducing semantic branch~\cite{He_2018_CVPR}, region proposals~\cite{li2018high}, hard negative mining~\cite{zhu2018distractor}, ensembling~\cite{he2018towards}, deeper backbone~\cite{li2018siamrpn++} or high-fidelity object representations~\cite{wang2018fast}.

Under the assumption that objects are under minor displacement and size change in consecutive frames, most modern trackers, including the ones mentioned above, use a steady search region, which is centered on the last estimated position of the target with the same ratio. Although it is very straightforward, this oversimplified prior often fails under occlusion, motion change, size change, or camera motion, as it is evident in the examples of Fig.~\ref{fig:teaser}. This motivates us to propose a robust motion prediction module that fits all these methods. 
\subsection{Video Forecasting}

The ability to predict and therefore to anticipate the future is an important attribute of intelligence. Many methods are proposed to improve the temporal stability of semantic video segmentation. Luc \etal~\cite{luc2017predicting} develop an auto-regressive convolutional neural network that learns to generate multiple future frames iteratively. Similarly, Walker \etal~\cite{walker2017pose} use a VAE to model the possible future movements of humans in the pose space. Instead of generating future states directly, many methods attempt to propagate segmentation from preceding input frames~\cite{jin2017video,nilsson2018semantic,jaderberg2015spatial}.

Unlike previous work, we extract a motion model for each object and set up a new search region for detection and segmentation accordingly.

\section{Method}

\begin{figure*}
    \centering
    \includegraphics[width=\textwidth]{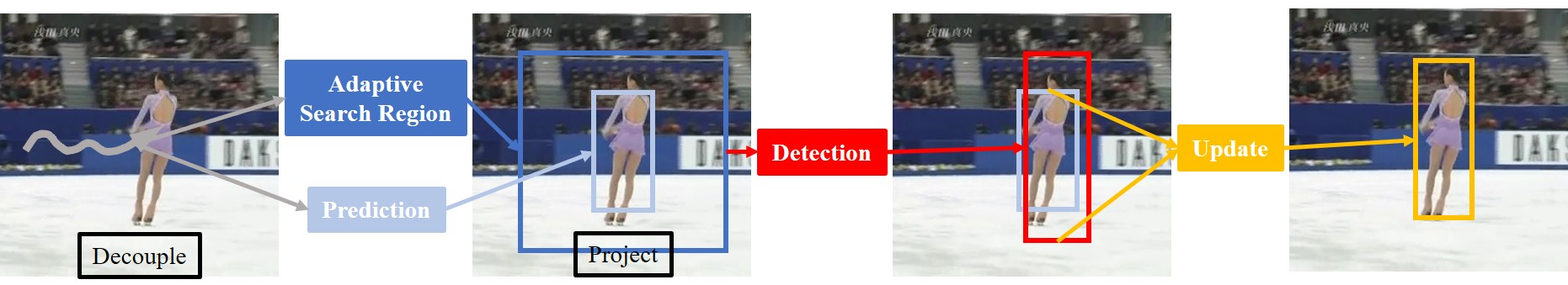}
    \caption{Our method first decouple camera motion and object motion. Second, we predict the object state in the future frame and create an adaptive search region for the detector to process. We then project the predicted state and search region back to the camera coordinate. Finally, we update the object state based on the measurement from the off-the-shelf object detector. State prediction and update are based on Kalman filter.}
    \label{fig:pipeline}
\end{figure*}

Our method first decouples object motion from camera motion. Second, we predict the object state in the future frame and create an adaptive search region for the detector to process. We then project the predicted state and search region back to the camera coordinate. We finally update the object state based on the measurement from the off-the-shelf object detector. State prediction and update are based on Kalman filter. We illustrate our framework in Fig.~\ref{fig:pipeline}.

\subsection{Decouple}\label{sec:method:Decouple}

Object motion in a given image is the superposition of camera motion and object motion. These motions may lie in different modes (\eg, random camera shaking, or object moving direction sudden change). Thus, predicting object motion in camera coordinates for a long horizon will lead to instability. To solve this problem, we first pick a reference frame ($F_{k}$, $k$ denotes $k^{th}$ reference frame) every $n$ frames and thus separate the long video into several pieces of short n-frame videos.

Second, we adopt the method proposed by ARIT~\cite{6751553} to decouple the camera motion and object motion within each short video. ARIT assumes that pending detection frame ($F_{k+t}$) and its reference frame ($F_{k}$) are related by a homography ($H_{k,k+t}$). To estimate the homography, the first step is to find the correspondences between two frames. As mentioned in ARIT, we combine SURF features~\cite{willems2008efficient} and motion vectors from the optical flow to generate sufficient and complementary candidate matches, which is shown to be robust~\cite{gauglitz2011evaluation,6751553}. Here we use PWCNet~\cite{sun2018pwc} for dense flow generation.

As a homography matrix contains eight free variables, at least four background points pairs should be used. We calculate the least square solution of eq.~\ref{eq:7} and optimize it to obtain robust solution through RANSAC~\cite{fischler1981random}, where $p_{k}^{bp}$ and $p_{k+t}^{bp}$ denotes random selected background matching pairs in $F_{k}$ and $F_{k+t}$ using the above mentioned features. 

\begin{equation} \label{eq:7}
H_{k,k+t}\times p_{k}^{bp}=p_{k+t}^{bp}
\end{equation}

\begin{figure}
    \centering
    \includegraphics[width=0.48\textwidth]{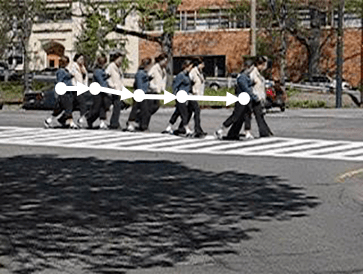}
    \caption{One example for decoupling camera motion and object motion (arrows illustrates the movement of object center).}
    \label{fig:decouple}
\end{figure}

Fig.~\ref{fig:decouple} illustrates the working principle of the decoupling step. The origin video for Fig.~\ref{fig:decouple} is a handheld video with trembling background. The motion of the pedestrians in the origin video is highly unpredictable with huge background uncertainties. However, by mapping the target frame towards the reference frame, the movement for pedestrians could be more predictable and continuous.

There are two cases where motion decouple is not applied: (i) no correspondences between the future frame and the reference frame (ii) outlier of RANSAC is larger than an error threshold. These always happen due to severe blur, where the tracking relies purely on target appearance.

For simplicity, the following calculations are under reference coordinate without further noticing.

\subsection{Prediction}\label{sec:method:Prediction}

We parameterize the bounding box as a set of five parameters, including the coordinate of the object center $(x, y)$ and object’s size $(w, h)$ and its confidence $c$. Our prediction does not need any training process and can be directly applied for inference. We formulate the state of object trajectory as a 6-dimensional vector $s = (x, y, w, h, v_x, v_y)$, where the additional variables $v_x, v_y$ represent the velocity of objects.

To predict the object state in the next frame, we use the dynamic model in the Kalman filter, which is shown in (eq.~\ref{predictor}). 

\begin{equation} 
    \label{predictor}
    {s}_{t|t-1} = F s_{t-1|t-1} + B u_t
\end{equation}
where ${s}_{t|t-1}$ is the prior state estimation given observations up to time $t-1$, $s_{t-1|t-1}$ is the optimal result of the previous state, $u_t$ is the control amount of the process at time t (zero in our case). And $F$ is the transition matrix, $B$ is the system parameter. In this paper, we approximate the inter-frame displacement of objects using the constant velocity model, which is initialized to zero for each object. 
Then we predict the covariance corresponding to the process result:
\begin{equation} 
    \label{pri_cov}
    {V}_{t|t-1} = F V_{t-1|t-1} F^T + Q
\end{equation}
where ${V}_{t|t-1}$ is a prediction of the covariance corresponding to the state $s_{t|t-1}$, $V_{t-1|t-1}$ is the covariance corresponding to the previous state $s_{t-1|t-1}$. And $Q$ is the covariance matrix of system noise (assumed to be Gaussian).

\subsection{Adaptive Search Region}\label{sec:method:search}

To alleviate the information needed to be processed by detectors and better filter out distractors, we dynamically set up a new search region in the coming frame centered at the predicted object position. The adaptive search region is modified with respect to the predicted velocity and object size. 

Given the estimated position, we setup the search region (a $S \times S$ square) as following:

\begin{equation}
S=k\sqrt{(w+p)(h+p)}
\end{equation}

\begin{equation}
    \label{parameter}
    k = 1 + 2\times sigmoid(||v||_2-\theta_v)
\end{equation}
where $p=\dfrac{w + h}{2}$, $v$ is the predicted velocity and $\theta_v$ is a pre-defined threshold.

\subsection{Project}\label{sec:method:Project}

We then project the estimated adaptive search region and predicted state back to the future frame as following: 
\begin{equation}
H_{k,k+t}\times P_{k}=P_{k+t} 
\end{equation}
where $P_{k}$ and $P_{k+t}$ are the key-points (centers and corners) in frame k and frame k+t. Since affine transformations do not respect lengths and angles, we recalculate object sizes based on projected corners.

\subsection{Detection}\label{sec:method:Detection}

It is worth noticing that our method does not depend on specific detection or segmentation methods. In this paper, we adopt SiamRPN++~\cite{li2018siamrpn++} for detection and SiamMask~\cite{wang2018fast} for segmentation, since they achieve a good balance between accuracy and speed. We refer readers to ~\cite{ren2015faster, li2018high} for understanding the region proposal branch and ~\cite{pinheiro2015learning,pinheiro2016learning} for understanding the mask branch.

\subsection{Update}\label{sec:method:Update}

To account for the uncertainty in prediction, we update the entire state space of trajectory based on its corresponding measurement, i.e., the detection result, and obtain the final trajectories using the following equation: 

\begin{equation}
    \label{post_predictor}
    s_{t|t} = s_{t|t-1} + K_t (D_t - O s_{t|t-1})
\end{equation}
where $D_t$ is the measurement (detection result in our case). And $O$ is the observation matrix, $K_t$ is the optimal Kalman gain defined by eq.~\ref{Kalman gain}. In our case, velocity states are not observable.

\begin{equation}
    \label{Kalman gain}
    K_t = \frac{V_{t|t-1} O^T}{O V_{t|t-1} O^T + R}
\end{equation}
where R is the covariance matrix corresponding to the measurement noise (assumed to be Gaussian).

We then perform the covariance update as following:

\begin{equation}
    \label{pos_cov}
    V_{t|t} = (I - K_t O)V_{t|t-1}
\end{equation}
where $I$ is an identity matrix.

We refer reader to ~\cite{10.1007/3-540-55426-2_49, 489052} for more details on Kalman filter. We only execute the aforementioned update when the detection confidence score is larger than a threshold $\theta_d$. If the detection confidence score is less than $\theta_d$, we update object states use only eq.~\ref{predictor} and eq.~\ref{pri_cov}. This can help to track objects under occlusions and large appearance changes. 

The motion consistency between video frames in different sliced videos with different reference frames could be an issue because the initialization of the velocity for the reference frame could be critical to the accuracy of the position update. To maintain the motion consistency, we choose the $n_{th}$ frame, which is the last frame in the sliced video, as the next reference frame with the refined position and velocity estimation from Kalman filter based on the former reference frame. Therefore, the velocity of the object, with respect to the new frame, could be initialized by mapping the refined velocity towards the new reference.

\section{Experiments}
\label{experiments}

In this section, we evaluate our approach on three tasks: motion prediction, visual object tracking (VOT-2016 and VOT-2018), and semi-supervised video object segmentation (DAVIS-2016 and DAVIS-2017). 

\subsection{Evaluation of motion prediction}

\paragraph{Datasets and settings}

We use VOT-2016~\cite{10.1007/978-3-319-48881-3_54} and VOT-2018~\cite{Kristan2018a} to evaluate the performance of motion prediction. Both datasets contain 60 public sequences with different challenging factors: camera motion, object motion change, object size change,  occlusion, and illumination change, which makes it extremely challenging for object motion prediction~\cite{Kristan2018a}. We use SiamMask~\cite{wang2018fast} for detection, which returns a segmentation mask for each tracking object. We thus use the center of mass of predicted mask as detected object position $(x, y)$. Our prediction of position and velocity is calculated as mentioned in Section~\ref{sec:method:Prediction}. For the baseline, the predicted position of the next frame (t+1) is always the same as the current frame (t), while object velocity is always predicted as 0. The ground truth position is set as the center of the annotated rotated bounding box, while the velocity is the difference between two consecutive positions. We evaluate the position error from ground truth with Euclidean distance and velocity error with Euclidean distance, cosine distance, and magnitude distance. Cosine distance is the cosine value between predicted velocity and ground truth velocity (the higher, the better). Magnitude distance is the absolute difference between the absolute value of predicted velocity and ground truth velocity. We adopt the reinitialize mechanism as used in the official VOT toolkit. When the segmentation has no overlap with ground truth, we reinitialize the tracking method with ground truth after five frames.

\paragraph{Results on VOT-2016 and VOT-2018}

\begin{figure}
\begin{center}
   \includegraphics[width=0.4\textwidth]{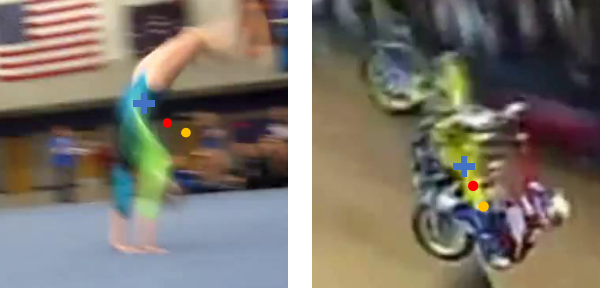}
  \caption{Position predictions (red for Ours, yellow for Baseline and blue cross for ground truth) (better view with color)}
  \label{fig:prediction error}   
\end{center}
\end{figure}

Table.\ref{prediction error} presents the comparison of position prediction error using the baseline method and our method. As it is shown in the table, for both of these two datasets, our method could dramatically reduce the prediction errors of the object position. The mean square error for object position on VOT-2018 could be reduced by half from 16 pixels to 8 pixels. Meanwhile, Fig.\ref{fig:prediction error} shows when the object velocity is high, our method could provide a more accurate prediction compared with Baseline, which does not consider the influence of object motion. The results prove that the decoupling strategy could reduce the background uncertainty, and the Kalman filter would provide a relatively reliable prediction for object position in the next frame. Higher accuracy for object position prediction could benefit the generation of search regions for object tracking and eventually improve the performance of object segmentation.

\begin{table}
\begin{center}
\begin{tabular}{c|c|c}
\hline
Dataset & Tracker &   Pos Err. \\ \hline  
\multirow{2}{*}{VOT-2016} & Baseline & 16.281 \\ \cline{2-3}
& Ours       & 8.198 \\ \hline
\multirow{2}{*}{VOT-2018} & Baseline       & 14.593 \\ \cline{2-3}
& Ours      & 8.744 \\ \hline
\end{tabular}
\caption{Position prediction error on VOT-2016 and VOT-2018}
\label{prediction error}
\end{center}
\end{table}

\begin{figure}[ht]
\begin{center}
   \includegraphics[width=0.4 \textwidth]{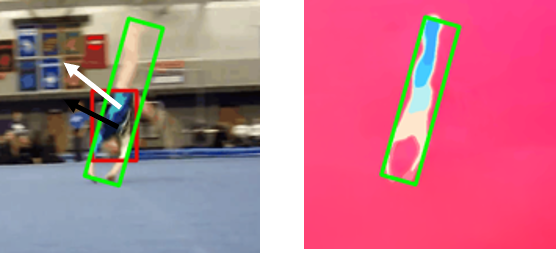}
  \caption{Velocity predictions (Left) (white for goundtruth, black for prediction, both extended by 5 times longer for better visualization) Optical Flow (Right) (better view with color)}
  \label{fig:velocity error}   
\end{center}
\end{figure}

For velocity, as can be seen in Table~\ref{velocity error}, our method significantly reduce the estimation error. In VOT-2018, Ours achieves 0.763 cosine distance, which is about 37-degree divergence from ground truth velocity direction. The main cause of the error is that objects are not always rigid, thus "center of mass" can approximate the overall motion of the object (Fig.~\ref{fig:velocity error}). The size change of objects will further increase the prediction error. However, with the correction procedure of the Kalman filter, this error (noise) can be stabilized. One possible solution to decrease velocity prediction error is tracking each part of non-rigid objects and grouping all parts together to get the final prediction~\cite{ramanan2007tracking}.

\begin{table}
\begin{center}
\begin{tabular}{c|c|c|c|c}
\hline
Dataset & Tracker & MSE Err. & Cosine & Mag.\\ \hline  
\multirow{2}{*}{VOT-2016} & Baseline & 8.274 & - & 8.274\\ \cline{2-5}
& Ours       &   4.596 & 0.667 & 3.190 \\ \hline
\multirow{2}{*}{VOT-2018} & Baseline &   7.006 & - & 7.006\\ \cline{2-5}
& Ours      &    4.298 & 0.793 & 2.929 \\ \hline

\end{tabular}
\caption{Velocity prediction error on VOT-2016 and VOT-2018}
\label{velocity error}
\end{center}
\end{table}

\subsection{Evaluation of VOT}

We adopt two widely used benchmarks for the evaluation of the object tracking task: VOT-2016 and VOT-2018. Here we adopted SiamRPN++ as our detection module. We compare our method against the state-of-the-arts using the official metric: Expected Average Overlap (EAO), which considers both accuracy and robustness of a tracker~\cite{Kristan2018a}. We further conduct an experiment on VOT-2018 for evaluating the performance under different conditions. 

\paragraph{Results on VOT-2016}

Table~\ref{VOT-2016} presents comparisons of tracking performance between our method and SiamRPN++ on VOT-2016 dataset. Our method improves the robustness by 23.2\%, and provide a 7.0\% gain of EAO, which achieves 0.505. The baseline is the state-of-the-art, other methods are not compared for simplicity.

\begin{table}
\begin{center}
\begin{tabular}{c|c|c|c}
\hline
           & \multicolumn{3}{c}{VOT-2016} \\ \hline
Trackers   & A        & R       & EAO     \\ \hline
SiamRPN++   & 0.633    & 0.181   & 0.472   \\ \hline
Ours & 0.642    & 0.139   & 0.505   \\ \hline
\end{tabular}
\end{center}
\caption{Comparison with SiamRPN++ on VOT-2016 }
\label{VOT-2016} 
\end{table}

\paragraph{Results on VOT-2018}

In Table~\ref{all result} we compare our method against eleven recently published state-of-the-art trackers on the VOT-2018 benchmark (A stands for accuracy and R stands for robustness). We establish a new state-of-the-art tracker with 0.431 EAO and 0.607 accuracy. In particular, our model outperforms all existing Correlation Filter-based trackers. This is very easy to understand since our baseline SiamRPN++ relies on deeper feature extraction, which is much richer than all existing Correlation Filter-based methods. Interestingly, our method even outperforms the baseline method. Previous research shows Siamese based trackers have strong center bias despite the appearances of test targets~\cite{li2018siamrpn++}. Thus, by estimating the center of the search region more accurately, Siamese trackers can also achieve better regression result (\eg, bounding box detection, or object segmentation). Besides, our method achieves the lowest robustness among all Siamese based trackers. This is even exhilarating because one of the key vulnerability of Siamese based trackers is the low robustness. The main reason is that most Siamese networks can only discriminate foreground from the non-semantic background~\cite{zhu2018distractor} and thus suffer from distinguishing target object from surrounding objects. Our proposed motion prediction module adopts a straightforward strategy and shows great improvement of robustness from 0.241 to 0.203, which provides another strategy to achieve better robustness: by setting more accurate and targeted search region. And our proposed modules only decrease the running speed by a small margin (5FPS) since all calculations are
done in GPU (RTX2080Ti).
\begin{table*}
\begin{center}
\resizebox{\textwidth}{!}{\begin{tabular}{c|cccccccccccc}
 & DaSiamRPN & SA\_Siam\_R & CPT & DeepSTSRCF & DRT & RCO & UPDT & SiamMask & SiamRPN & MFT & SiamRPN++ & Ours \\ \hline
EAO $\uparrow$ & 0.326 & 0.337 & 0.339 & 0.345 & 0.356 & 0.376 & 0.378 & 0.380 & 0.383 & 0.385 & 0.410 & \textbf{0.431} \\
Accuracy $\uparrow$ & 0.569 & 0.566 & 0.506 & 0.523 & 0.519 & 0.507 & 0.536 & 0.609 & 0.586 & 0.505 & 0.594 & 0.607 \\
Robustness $\downarrow$ & 0.337 & 0.258 & 0.239 & 0.215 & 0.201 & 0.155 & 0.184 & 0.276 & 0.276 & 0.140 & 0.241 & 0.203 \\
Speed (FPS) $\uparrow$ & 160 & 32 & 14 & 24 & $< 1$ & 7 & $< 1$ & 56 & 200 & $< 1$ & 35 & 30 \\ \hline
\end{tabular}}
\caption{Comparison with the state-of-the-art trackers under EAO, Accuracy, Robustness and speed on the VOT-2018 dataset.}
\label{all result} 
\end{center}
\end{table*}

\begin{figure*}[ht!]
\begin{center}
   \includegraphics[width=\textwidth]{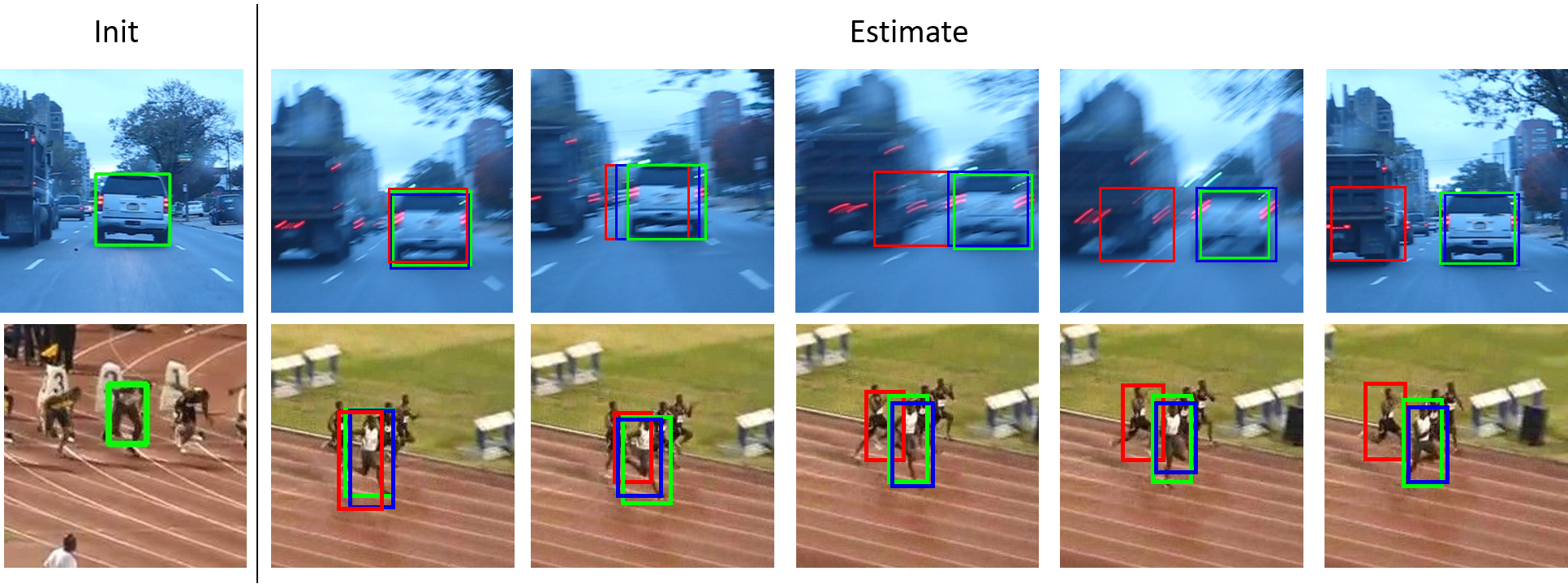}
  \caption{Qualitative result of SiamRPN++ and our method : green box is the ground truth, red box is the bounding box from SiamRPN++, and blue box is our method.}
  \label{Qualitative result}   
\end{center}
\end{figure*}

To further analysis where the improvements come from, we show the qualitative results of our method and the baseline SiamRPN++ (Fig.~\ref{Qualitative result}). Just as mentioned above, the robustness comes from less tracking object switching and missing. For example, as for the car scenario in Fig.~\ref{Qualitative result}, when the camera shakes, the center of the search region of SiamRPN++ will shift to the left of the tracking car, and finally catches the truck. On the contrary, the center of our search region stays on the tracking car, since our model considers camera motion. This stability comes from the decoupling of camera motion. Another example is Bolt, the second row in Fig.~\ref{Qualitative result}. When Bolt accelerates, SiamRPN++ will be easily distracted by other runners, but our model does not fail because it considers the speed of Bolt. This stability comes from object velocity estimation. These unique features contribute to the performance of our method under large camera motion, fast object motion, and occlusion. In short, by predicting object position accurately, our model can focus on a more targeted search region and thus achieve better detection and segmentation performance. 

\begin{table}
\begin{center}
\begin{tabular}{c|c|c|c} \hline
Datasets & Methods & J & F \\ \hline
\multirow{2}{*}{Davis-2016} & SiamMask & 0.713 & 0.674 \\  \cline{2-4}
& Ours & 0.732 & 0.692 \\ \hline
\multirow{2}{*}{Davis-2017} & SiamMask & 0.543 & 0.585 \\  \cline{2-4}
& Ours & 0.554 & 0.604 \\  \hline
\end{tabular}
\caption{J and F Results on Davis-2016 and Davis-2017 }
\label{Davis} 
\end{center}
\end{table}

\begin{figure*}[ht!]
\begin{center}
   \includegraphics[width=\textwidth]{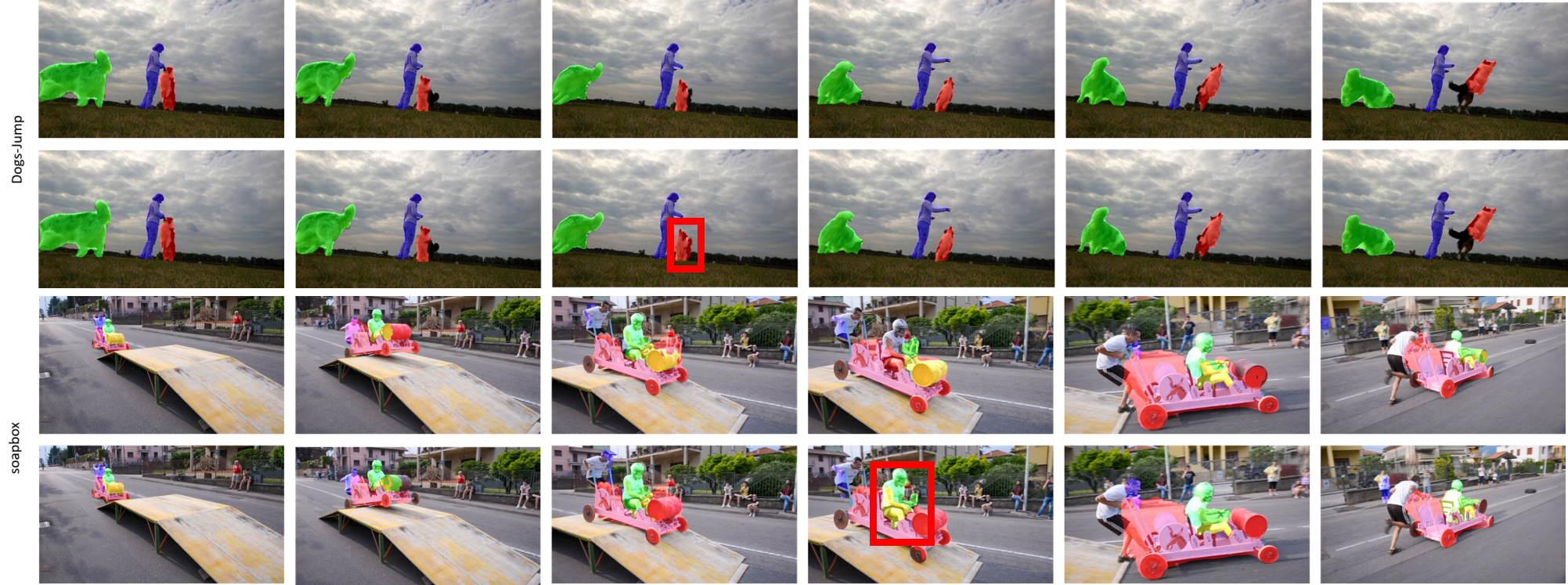}
  \caption{Qualitative result of SiamMask and our method on DAVIS: First row and third row are the results from SiamMask. Second row and fourth row are the results from same videos using our method. (better view with color)}
  \label{fig:davis}   
\end{center}
\end{figure*}

\subsection{Evaluation for VOS}

Although this paper focuses on VOT, we also test our method on VOS. 

\paragraph{Datasets and settings}

We also report the performance of our method on standard VOS datasets DAVIS-2016~\cite{perazzi2016benchmark} and DAVIS-2017~\cite{pont20172017}. For both datasets, we use the official performance measures: the Jaccard index (J) to express region similarity and the F-measure (F) to express contour accuracy. We use SiamMask as our segmentation module and adopt the semi-supervised setup. We fit bounding boxes to object masks in the first frame and use these bounding boxes to initialize our tracker. 

\paragraph{Results on DAVIS-2016 and DAVIS-2017}

Table~\ref{Davis} presents the comparison of VOS results using SiamMask and our proposed motion prediction model on Davis-2016 and Davis-2017 datasets. The effectiveness of our approach is limited on Davis-2016 and Davis-2017 datasets. The main reason is that DAVIS datasets have less camera motion or fast object motion. However, segmentation can still benefit from more accurately cropped search region. \eg, The dog in the third frame of the "Dogs-Jump" video is segmented more completely through motion prediction. However, SiamMask misses the tail of the same dog during segmentation. Another example is the person in the fourth frame of the "Soap-Box" video. Our method separates this person from the soapbox. However, SiamMask mixes its segmentation with the surrounding pixels. Further, SiamMask fails to distinguish the person mask from the drum of the soapbox because the drum occupies the previous position of the person, which can not be handled without motion assumption. Though our pre-tracking procedure, our method can separate specific instance from its neighboring instance and thus get a more accurate segmentation. We show that our proposed method does a better job at segmenting under crowded scenarios. 
For more qualitative results, please refer to Fig.~\ref{fig:davis}. 

\subsection{Ablation studies}

Table~\ref{ablation} compares the contribution of each module in our pipeline. Based on VOT-2018 dataset, we evaluate motion decouple (MD), motion prediction (MP), and adaptive search region (ASR) with the baseline approach (SiamMask). It can be observed from Table~\ref{ablation} that the motion decouple and motion prediction play important roles in our method. The adaptive search region module only contributes 0.02 EAO improvement and using adaptive search region only even decrease the performance. This is because without motion decouple, the motion velocity might be very noisy. However, as we can see from Table~\ref{ablation}, both of motion prediction and adaptive search region have the potential to improve accuracy with correct motion decouple.

\begin{table}[]
\begin{center}
\begin{tabular}{c|ccc} \hline
 & EAO & A & R \\ \hline
SiamMask & 0.380 & 0.609 & 0.276 \\
SiamMask + ASR & 0.379 & 0.604 & 0.280 \\
SiamMask + MD + ASR & 0.382 & 0.610 & 0.268 \\
SiamMask + MP & 0.384 & 0.610 & 0.262 \\
SiamMask + MD + MP & 0.394 & 0.611 & 0.234 \\
Ours & 0.397 & 0.612 & 0.220 \\ \hline
\end{tabular}
\caption{Ablation studies for motion prediction and adaptive search region on VOT-2018 dataset.}
\label{ablation} 
\end{center}
\end{table}

\section{Conclusion}

In conclusion, we show that motion prediction can still play an important role in visual object tracking. We propose a method that combines motion prediction with off-the-shelf appearance-based trackers. Although our baseline system is a straight forward combination of standard methods, we obtain the state-of-the-art results on VOT. We also show consistent improvements on VOS. We hope our work can inspire more studies in considering the relationship between appearance and motion information in modern trackers. 

\nocite{8885685, Wang_2020_WACV}
{\small
\bibliographystyle{IEEEtran}
\bibliography{egbib}
}

\end{document}